\PassOptionsToPackage{unicode=true}{hyperref} 
\PassOptionsToPackage{hyphens}{url}
\documentclass[11pt,a4paper]{article}
\usepackage[hyperref]{naaclhlt2019}
\usepackage{times}
\usepackage{graphicx}
\usepackage{amssymb,amsmath}
\usepackage{bm}

\usepackage{url}
\aclfinalcopy 

\usepackage{tablefootnote}
\usepackage[utf8x]{inputenc}

\usepackage[]{natbib}

\title{Geolocating Political Events in Text}
\author{Andrew Halterman \\
       Department of Political Science \\
       Massachusetts Institute of Technology \\
       {\tt ahalt@mit.edu}}
       
\date{}

\begin{document}

\maketitle

\begin{abstract}
This work introduces a general method for automatically finding the
locations where political events in text occurred. Using a novel set of
8,000 labeled sentences, I create a method to link automatically
extracted events and locations in text. The model achieves human level
performance on the annotation task and outperforms previous event
geolocation systems. It can be applied to most event extraction systems
across geographic contexts. I formalize the event--location linking
task, describe the neural network model, describe the potential uses of
such a system in political science, and demonstrate a workflow to answer
an open question on the role of conventional military offensives in
causing civilian casualties in the Syrian civil war.
\end{abstract}

\section{Introduction}\label{introduction}

Researchers in social science, and especially in comparative politics
and security studies, are increasingly turning toward micro-level data,
with subnational variation at very fine resolutions becoming a major
source of empirical puzzles and evidence in these fields. At the same
time, text data is becoming one of the most important sources of new
data in social science. I develop and describe a method that enables
researchers to connect these two trends, automatically linking events
extracted from text to the specific locations where they are reported to
occur.

Specifically, I develop a method that, given a sentence and an event's
verb in the sentence, will return the place names from the sentence
where the event took place. Formulated as a general task, this is an
unsolved problem in both political science and computer science. Drawing
on a set of 8,000 hand-labeled sentences, I train a recurrent neural
network that draws on a rich set of linguistic features to label a
sequence of text with labels for whether the word is a location word
corresponding to a specified verb. Measured by token, the model produces
precision and recall scores of over 0.83, compared with a rule-based
model's 0.25--0.29. A software implementation and example workflow is
provided.

I provide an example application, creating a new dataset on the
locations of military offensives in Syria and contributing to an ongoing
debate in conflict studies on the causes of civilian casualties in civil
war. The model is general enough for applied researchers to use in other
contexts, including the study of protests, political mobilization,
political violence, and electoral politics. The new shared dataset will
enable other researchers in NLP to contribute to this task and the wider
research project of better extracting political relationships from text.

\section{Task and Formulation}\label{task-and-formulation}

Event--location linking sits within a larger set of techniques for
extracting information on political events from text, including entity
extraction and toponym resolution.

Event extraction is the process of recognizing defined event types in
text (e.g. ``attack'' or ``protest'') and extracting and classifying the
actors involved in the events. Many approaches to this task exist in
both political science and NLP, using both rule-based and machine
learning coders (Schrodt, Davis, and Weddle 1994; O'Connor, Stewart, and
Smith 2013; Schrodt, Beieler, and Idris 2014; Boschee et al. 2015;
Beieler 2016; Beieler et al. 2016; Hanna 2017; Keith et al. 2017).

To be useful in subnational research, these events require information
on the location where they occurred. A second related information
extraction task is ``geoparsing'', the process of recognizing place
names in text (``toponym recognition'') and resolving them to their
coordinates or gazetteer entry (``toponym resolution''). Some work on
geoparsing, also referred to as ``georeferencing'' or ``toponym
resolution'' exists (Leidner 2008; Hill 2009; Speriosu and Baldridge
2013; Berico Technologies, n.d.; D'Ignazio et al. 2014; Gritta et al.
2017; Halterman 2017; Avvenuti et al. 2018). Performing this task
requires disambiguating place names using heuristics or a model (in a
particular document, is ``Prague'' the capital of the Czech Republic or
the town in Oklahoma?).

The task that this paper addresses sits between the two: given an
extracted event in a sentence, which of place name is the location where
the event occurred? Consider the following sentence as a running
example:

\begin{quote}
After establishing a foothold in the northern Aleppo towns of Tadif and
Al-Bab, the Turkish Army and allied Syrian rebels launched an offensive
on its neighbouring town of Bza'a, a spokesperson for Ankara said today.
\end{quote}

An event extraction system may identify events such an ``establish
foothold'' event or a ``launch offensive'' event. A geoparser would be
concerned with recognizing the place names in the text (``Aleppo'',
``al-Bab'', ``Bza'a'') and resolving them to their correct coordinates
(made difficult by ``Aleppo'' being the governorate here, not the city).
An event--location linking system of the kind introduced here would
associate the ``establish foothold'' event with ``al-Bab'' and the
``launch offensive'' event with ``Bza'a''.

The task can be formalized as follows. Consider
\(\bm{X} = \{w_1,...w_n\},\) a sentence of \(n\) tokens. Given an event
\(e_k\), the location where event \(e_k\) occurred is defined as a set
of tokens \(\bm{G}_k = \{g_1,...,g_j\}\). For \(e_1 =\) ``establish a
foothold'', \(G_1 = \{\text{Tadif, Al-Bab}\}\).

Because a sentence can contain multiple events, the set of event
locations \(\bm{G}_k\) and \(\bm{G}_{k'}\) are not equivalent for
\(k \neq k'\). For \(e_2 =\) ``launch an offensive'',
\(\bm{G}_2 = \text{\{Bza'a\}}\). \(\bm{G}_k\) can have zero elements,
one, or several elements. Thus, for
\(e_3 = \text{``said"}, \bm{G}_3 = \{\}\), as the ``said'' event is not
associated with a specific place.

Each token \(w_i \in \bm{X}\) is given a label \(y_{i}^{(k)}\), where

\[y_{i}^{(k)} = \begin{cases} 
1 \text{ if } w_i \text{ is where event } k \text{ occurred} \\
0 \text{ otherwise}
\end{cases}\]

To make the estimation of \(\bm{\hat{y}}^{(k)}\) tractable, I make
several assumptions.

First, in order to condition on the event \(e_k\), I assume that the
information provided by the verb \(v_k\) of the event \(e_k\) is
sufficient.\footnote{By ``verb'' I mean the highest verb on the
  dependency tree that is uniquely part of event \(e_k\). In
  dictionary-based event coding methods, this is in practice the lexical
  ``trigger'' word for the event, though the event--location linking
  method is agnostic to how the event is coded.} Thus,
\(\bm{y}^{(k)} = f(\bm{X}, e_k) := f(\bm{X}, v_k)\). This assumption,
that events are ``anchored'' by a verb, is a common assumption in
semantic role labeling, a closely related task to event--location
linking (Palmer, Gildea, and Kingsbury 2005; White et al. 2016;
Marcheggiani and Titov 2017).\footnote{Though consider the phrase
  ``After the riots in Gujarat\ldots{}''. This sentence reports a
  ``riot'' event but without a verb. These clausal mentions of events
  are rarely coded by event extraction systems, both because of
  difficulty in coding and because they often describe historical,
  rather than contemporary events, meaning the decision to require a
  verb has little practical difference.}

Second, I assume that an adequate representation of each word \(w_i\) is
\(\phi(w_i)\), where \(\phi\) is a feature-making function that maps
\(w_i\) from a high dimensional one-hot vector to a lower dimensional
dense encoding, drawing on the context of the word in the sentence.
Applied to a sentence,

\[\Phi(X) = \{\phi(w_1), \ldots, \phi(w_n)\}.\].

Thus,

\[\bm{\hat{y}^{(k)}} = \hat{f}(\Phi(\bm{X}), v_k).\]

Finally, I assume that the event location status \(y_i^{(k)}\) of word
\(w_i\) is conditionally independent of other words' labels
\(y_{j \neq i}^{(k)}\) after conditioning on the matrix of sentence
context \(\Phi(X)\). Making this assumption greatly simplifies
estimation, as the task of assigning labels can be decomposed into a set
of independent tasks:

\[\begin{aligned}
\bm{\hat{y}^{(k)}} &=  \hat{f}(\Phi(\bm{X}), v_k) \\
&= \{\hat{f}(\phi(w_1), v_k),...,\hat{f}(\phi(w_n), v_k) \}
\end{aligned}
\]

This assumption only carries costs if words' labels affect each other
through a mechanism outside of \(\bm{X}\). The assumption seems
warranted here, though, because of the binary nature of the
classification task.\footnote{This conditional independence of labels
  assumption is generally not made in part of speech tagging, dependency
  parsing, or named entity recognition. In these tasks, each word can be
  assigned one of many possible labels, and past labels dramatically
  change label probabilities. (For example, if a word is predicted to
  have the part-of-speech label \textsc{verb}, the following word cannot
  be labeled be \textsc{verb} if the sentence is to be grammatical).
  These tasks required more sophisticated beam search or shift-reduce
  models (Goldberg 2017; Jurafsky and Martin 2018).}

\section{Previous work}\label{previous-work}

Many existing open source geolocated event datasets, including GDELT and
Phoenix, make no effort to explicitly link events and locations, simply
returning a top location from a sentence, without using information on
the extracted event to inform the geolocation step, which has also been
used in NLP (Aone and Ramos-Santacruz 2000).\footnote{ICEWS uses a
  proprietary system to link events and locations that is not documented
  or accessible to researchers (Lautenschlager, Starz, and Warfield
  2017).} Two recently proposed models do attempt to find events'
locations, however (Imani et al. 2017; Lee, Liu, and Ward 2018). Both
make a major simplifying assumption, that returning the correct location
does not depend on conditioning on an event of interest:
\(\bm{G}_k = \bm{G}_{k'} \text{ for all } e_k, e_k'\). The advantage of
this assumption is that each model can use a simple bag-of-words model
that does not account for word order or grammatical information, but it
means that the labels they produce for text with multiple events and
locations will be incorrect for at least some events.

Imani et al. (2017) propose a method for finding the ``primary focus
location'' of a story, which they define as ``the place of occurrence of
the event'' (1956). Their method makes the simplifying assumption that
documents have one single, fixed ``focus location'' that is invariant to
different potential events in the document. During training and testing,
they eliminate all documents with multiple events and multiple ``focus
locations.'' Their model discards word order information, representing
each sentence as a weighted average of pretrained word embedding, and
use this feature vector as an input to an SVM that predicts which
sentence contains the ``focus location.'' Then, the most frequent place
name in the ``focus sentence'' is the ``focus location.''

Lee, Liu, and Ward (2018) also make several other restrictive
assumptions. The implementation of their model is only able to located
events to the governorate/province (ADM1) level, and finds locations
based on a dictionary search of known place names: \(y_i = 0\) for any
\(w_i\) that is not present in the list of place names. This limits the
maximum accuracy to a relatively coarse level, and prevents the method
from recognizing places that are not on a relatively short list of place
names, which is unlikely to contain more rural or obscure places. Any
findings will be biased toward more populated areas, a known problem in
political violence research (Kalyvas 2004; Douglass and Harkness 2018).
Second, they learn a different \(f\) for different event types,
requiring documents to be classified into event types before
geolocation, requiring a training round with labeled data for each event
type and preventing parameters from being shared across models for
different event types.

Other work, in natural language processing, is related but not directly
applicable. Existing semantic role labeling and event extraction tasks
sometimes include location slots for events (e.g. Doddington et al.
2004), but none are precisely suited to a general system focused on
political events. FrameNet (Baker, Fillmore, and Lowe 1998) events have
highly specific slots for different event types, while PropBank (Palmer,
Gildea, and Kingsbury 2005) defines locations in a broad way that
includes non-tangible places (``keep in our \emph{thoughts}). A more
specific literature on spatial information in text also exists. For
instance, the SpaceEval task (Pustejovsky et al. 2015) provides a
comprehensive ontology of spatial relations in text. These relations are
focused on entities, rather than events, and provide more detail than is
desirable in a application-oriented model. The task as I have formulated
it thus seeks to be much more general, in that it attempts to locate any
type of event, but also more limited, in that it focuses solely on where
events occurred, rather than a larger set of spatial relations between
entities.

The closest existing work in NLP is Chung et al. (2017). They attempt to
find both explicit and implicit event locations in text, using a corpus
of 48 documents. They use a rule-based system built on top of word
embedding similarity and existing gold standard OntoNotes grammatical
information to infer the locations of events. While the system shows
good performance and is able to geolocate events even when the location
information is not provided directly in the sentence, it relies on
access to gold standard dependency parse information in a single domain
of text.

\section{Data}\label{data}

Implementing an automated procedure for geolocating events required
collecting a novel set of data. I created a new dataset of around 8,000
labeled sentences in English, each of which is annotated with an event
verb and its corresponding location or locations (if any).\footnote{The
  data and related materials are available at
  \url{https://github.com/ahalterman/event_location}} Sentences may have
multiple annotations corresponding to different verbs of interest.
Sentences were selected from a range of sources to maximize the
applicability of models trained on the data. The text is drawn from a
wide range of sources, including an assortment of international papers
and news wires (50\%), a selection of local English-language media from
Syria (35\%), and non-news sources such as Wikipedia, atrocity
monitoring reports, or press releases (15\%). Annotation consisted of
selecting a verb, either using a dictionary of specified verbs that
focused on territorial capture-type events, or using verbs automatically
detected using spaCy with the exception of ``to be'' to ensure the
generalizability of the data. The verbs were not filtered through an
event extraction system to keep the set as general as possible.
Annotators then selected the tokens representing the event locations for
the verb, if any. Around 5,000 annotations were provided by a research
assistant and 5,000 were annotated by me. After annotation, each
sentence looks something like the following:

\begin{quote}
He was speaking a day after Ankara {[}launched
\textsubscript{\textsc{verb}}{]} an offensive in the Syrian towns of
{[}Jarablus \textsubscript{\textsc{event\_loc}}{]} and {[}Kobane
\textsubscript{\textsc{event\_loc}}{]}.
\end{quote}

Annotations consist of the most specific named place or places, in
contrast to previous approaches that were limited to the city (Imani et
al. 2017) or the governorate/province (Lee, Liu, and Ward 2018). Events
can have no reported event, a single event with multiple location tokens
(``New York''), or multiple event locations (``New York and
Washington''). The modal number of locations is one (49\%), followed by
no locations (47\%), and multiple locations comprise the remainder
(3\%). Most locations consist of a single token (69\%), 19\% are two
tokens, and the remaining 12\% are three or more. Sentences have a large
number of verbs, and thus a potentially large number of events. The
average number of verbs per sentences is 3.6, after excluding auxiliary
verbs. Only 9\% contain a single non-auxiliary verb, and 21\% contain
five or more verbs.

\section{Model}\label{model}

I develop two neural network models to perform the event--location
linking task. I also describe a rule-based baseline model, along with
existing models from the literature as comparisons.

I use as a baseline model a rule-based event--location linker that
locates an event to the automatically recognized location word in
closest linear proximity to the event's verb. This model provides
per-event locations, unlike existing models, and incorporates a minimal
sentence distance feature.

Neural networks are now the dominant approach to most of natural
language processing (Goldberg 2017; Jurafsky and Martin 2018) so they
are the models adopted here. Determining the event locations in a
sentence using neural networks requires a language representation that
preserves word order and useful grammatical information in the sentence.
I preprocess the sentence by representing each word as a concatenation
of the following information generated by the spaCy NLP library
(Honnibal and Montani 2017) pretrained GloVe vector, dependency label,
named entity label, part-of-speech tag, an indicator for whether the
word is the event verb of interest, the (signed) distance between the
word and the indicated verb, and the distance between the verb and the
token on the dependency tree. I use the same features for two neural
network models. Both of the neural net models below look at a token,
along with its context, and make a binary prediction for whether the
token is an event location for the specified event.

The first neural network model uses a series of stacked convolutional
layers. Some research suggests that convolutional neural networks (CNNs)
perform equivalently to recurrent neural networks on sequence modeling
tasks with lower computational cost (Bai, Kolter, and Koltun 2018). Each
convolution looks at three inputs (words) at once, and slides down the
sentence one token at a time. By stacking layers on top of each other,
the elements of the final output of the final convolutional layer
includes information from across the sentence. I use residual layers (He
et al. 2016), which are now the state-of-the-art on image recognition
tasks. Residual layers help prevent the ``vanishing/exploding gradient''
problem that deep neural networks often encounter, and speed the model's
fitting. A CNN with residual layers empirically outperformed a model of
similar depth and structure without residual blocks, and is
theoretically justified because they allow me to train a deeper network
with lower demands on my limited pool of input data. After training and
evaluating several dozen models, the best performing CNN model used 7
residual layers with 64 hidden nodes in each, followed by two dense
layers with 512 nodes each with a dropout of 0.4 and ReLU activation.

\begin{figure*}
\centering
\includegraphics[width=1.6\columnwidth]{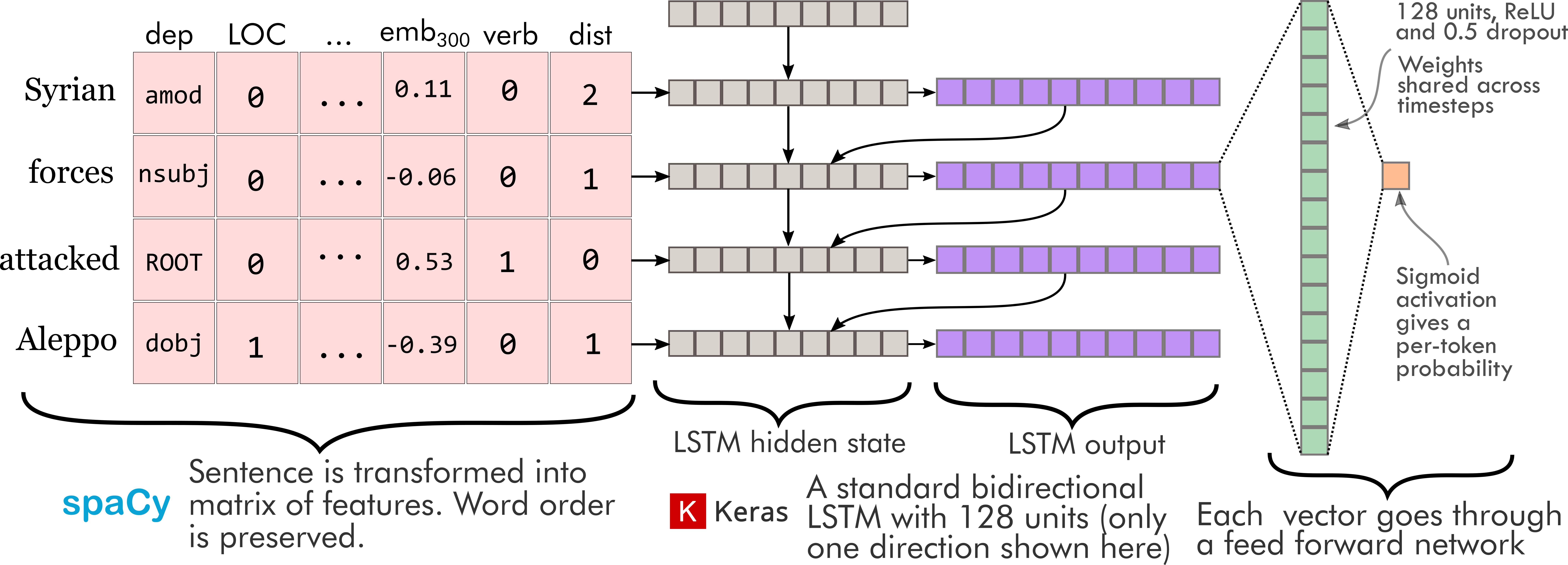}
\caption{High-level schematic of LSTM model}
\label{fig:lstm}
\end{figure*}

The second class models is recurrent neural network (RNN), specifically
a long short-term memory (LSTM) network (Hochreiter and Schmidhuber
1997), which explicitly models the sequential nature of their input data
(see Figure \ref{fig:lstm}). RNNs are the dominant approach to sequence
modeling tasks in natural language processing and achieve
state-of-the-art results on many tasks (Goldberg 2017). LSTMs store an
internal state at each step of the input data in the form of a hidden
vector. In contrast to vanilla RNNs, LSTMs can learn when to add
information from their current input step to the hidden state and when
to ``forget'' information from the hidden state. In theory, this allows
LSTMs to learn much longer relationships than they would otherwise be
able to. Bidirectional LSTMs are the standard extension to LSTMs when
the model has access to the ``future'', and compute two state vectors
for each input step: one from the left and one from the right. These two
vectors are concatenated and used as input to the rest of the model. The
best LSTM network I trained used a bidirectional LSTM with a hidden size
of 128 and 0.2 recurrent dropout, followed by a dense layer of 128 with
ReLu and 0.5 dropout, and a final binary output node for each time step.

All models were trained in Keras with a Tensorflow backend on a
multicore CPU.\footnote{The models are are available in Mordecai, an
  open-source document geoparser:
  \url{https://github.com/openeventdata/mordecai}}

In addition to my baseline and neural network models, I also perform
comparisons with three existing approaches. First, PropBank is included
as a point of comparison. The PropBank includes an ArgM-LOC label in
Palmer, Gildea, and Kingsbury (2005). The framing of the location task
in PropBank is quite different than the generalized event--location
linking task I introduce, as I describe above, but the performance of
the baseline model in Palmer, Gildea, and Kingsbury (2005) on the task
serves as another baseline. Second, I modify Profile (Imani et al. 2017)
to accept new text and compare its performance on my new labeled data.
Third, I report the best-case values from Chung et al. (2017).

Finally, I estimate the expected real-world performance of a human
annotator by comparing an annotator's performance to a ``gold standard''
set of annotations. To produce the gold set, I randomly selected
sentences annotated by the research assistant. I reannotated them,
skipping ambiguous sentences. Sentences with the same annotations in the
two periods were included in the gold evaluation set totalling 500
sentences. I could then compare RA performance with a ``gold'' measure
of performance.

\section{Evaluation}\label{evaluation}

I evaluate these and several existing models on the task and the
English-language dataset I introduce. To evaluate the performance of
each model, I assess accuracy on both a per-token and per-sentence
basis. For per-token accuracy, I take a common approach of calculating
the precision and recall in the evaluation sentences. Each model is
evaluated on how well it can can produce, for each token \(w_i \in X\)
whether \(w_i\) is an event location for event \(k\). This evaluation
approach allows ``partial credit'' for models that that may miss or
falsely include a single token and is a common approach to evaluating
sequence labeling tasks (Strötgen and Gertz 2016). I also include a
second measure that more closely matches real-word accuracy. This
measure reports the proportion of documents for which the annotation
produced by the method exactly matches the correct label for each token
in the document:
\(\hat{y}_{i}^{(k)} = y_{i}^{(k)} \forall i \in \bm{X}\). The results
for the word distance baseline measure, existing approaches, expected
human performance, and the two models I develop are reported in Table
1.\footnote{Results are not reported for the method developed by Lee,
  Liu, and Ward (2018). Unlike the other approaches, this method only
  geolocates to the province/ADM1 level, which is much coarser than
  these other techniques. It can only find place names on a provided
  whitelist of names, and models are customized to specific countries
  and events, making it unsuitable for this more general task of linking
  arbitrary locations and events. Finally, the replication code provided
  is not easily applicable to new datasets, only to run the initial
  experiments.}

\begin{table}
\begin{tabular}{lcccc}
\hline
\textbf{Model} & \textbf{Prec} & \textbf{Rec} &  \textbf{F1} & \textbf{Sentence}\tabularnewline
\hline
Baseline & 0.29 & 0.25 & 0.27 & 0.28 \tabularnewline
Profile & 0.54 & 0.29 & 0.37 & 0.51\tabularnewline
\emph{PropBank}\tablefootnote{These
  numbers are performance on the PropBank dataset, not on the dataset I
  create.} & \emph{0.61} & \emph{0.39} & \emph{0.47} & -\tabularnewline
CNN & 0.70 & 0.54 & 0.61 & - \tabularnewline
\emph{Chung et al.}\tablefootnote{Performance of Chung et al's model on their
corpus of 48 OntoNotes documents. The maximum values achieved for precision,
recall, and F1 across their models are reported here. Note that the results on my model report per-token
precision and recall, while they report per-location precision and recall.} & \emph{0.74} & \emph{0.62} &
\emph{0.62} & - \tabularnewline

Annotator & \textbf{0.88} & 0.65 & 0.74 & 0.73\tabularnewline
\textbf{LSTM} & 0.85 & \textbf{0.83} & \textbf{0.84} & \textbf{0.77}\tabularnewline
\hline
\end{tabular}
\caption{Per-token precision, recall, and F1 scores, and full-sentence accuracy for the word
distance baseline model, expected human performance, existing results from the
literature, and new model-based approaches.}
\end{table}

The word distance baseline model, which locates an event to the closest
recognized place name in the text, performs the worst of any model,
perhaps due to the unreliability of the distance heuristic itself,
errors in the NER system, and the model missing places when multiple
correct locations are present.

Profile (Imani et al. 2017) performs next worst, with an token-level F1
score of 0.37. The model is unable to to vary its location prediction by
event type, meaning that it will correctly locate at best one event's
location in a multi-event, multi-location sentence. Profile also returns
only one location per sentence, lowering its accuracy on events that
occur in multiple locations. Profile's intended use case is on longer
pieces of text: its poor performance on this task should be taken only
as an indication of its ability to geolocate events in text, not on its
ability to find the primary ``focus'' (D'Ignazio et al. 2014) location
of a piece of text.

PropBank is included as a point of comparison. The PropBank values are
reported for the ArgM-LOC label in Palmer, Gildea, and Kingsbury (2005).
The framing of the location task in PropBank is quite different than the
generalized event--location linking task I introduce, as I describe
above. The reported F1 score of 0.47 can be taken as a reasonable
baseline performance on an event--location linking task. Chung et al's
(2017) accuracy on their dataset and version of the task is the best of
any prior model.

The LSTM model performs much better than the CNN model, even after
extensive tuning for the CNN model. Inspection of the CNN model's output
(not included) indicates that the model seems to not learn long-distance
relationships well, and failed to appropriately change probability
weights when the verb of interest changed. The LSTM model, in contrast,
performs very well and is very sensitive to changes in the input verb:
the same sentence with two different flagged verbs of interest will
produce quite different results for those events' location. The LSTM and
CNN are comparable in training time.

Notably, the LSTM also outperforms an estimate of expected human
performance on the event--location linking task. While humans are able
to pick up on nuance and deal with grammatical complexity that machines
still cannot handle, humans are also unsuited to the tedium of labeling
thousands of sentences and may be susceptible to drift in their
definitions or understanding of the task. Not only is the automated
method vastly cheaper and faster than a human process, it does so with
accuracy at least as good.

\subsection{Ablation test}\label{ablation-test}

\begin{figure*}
\centering
\includegraphics[width=1.7\columnwidth]{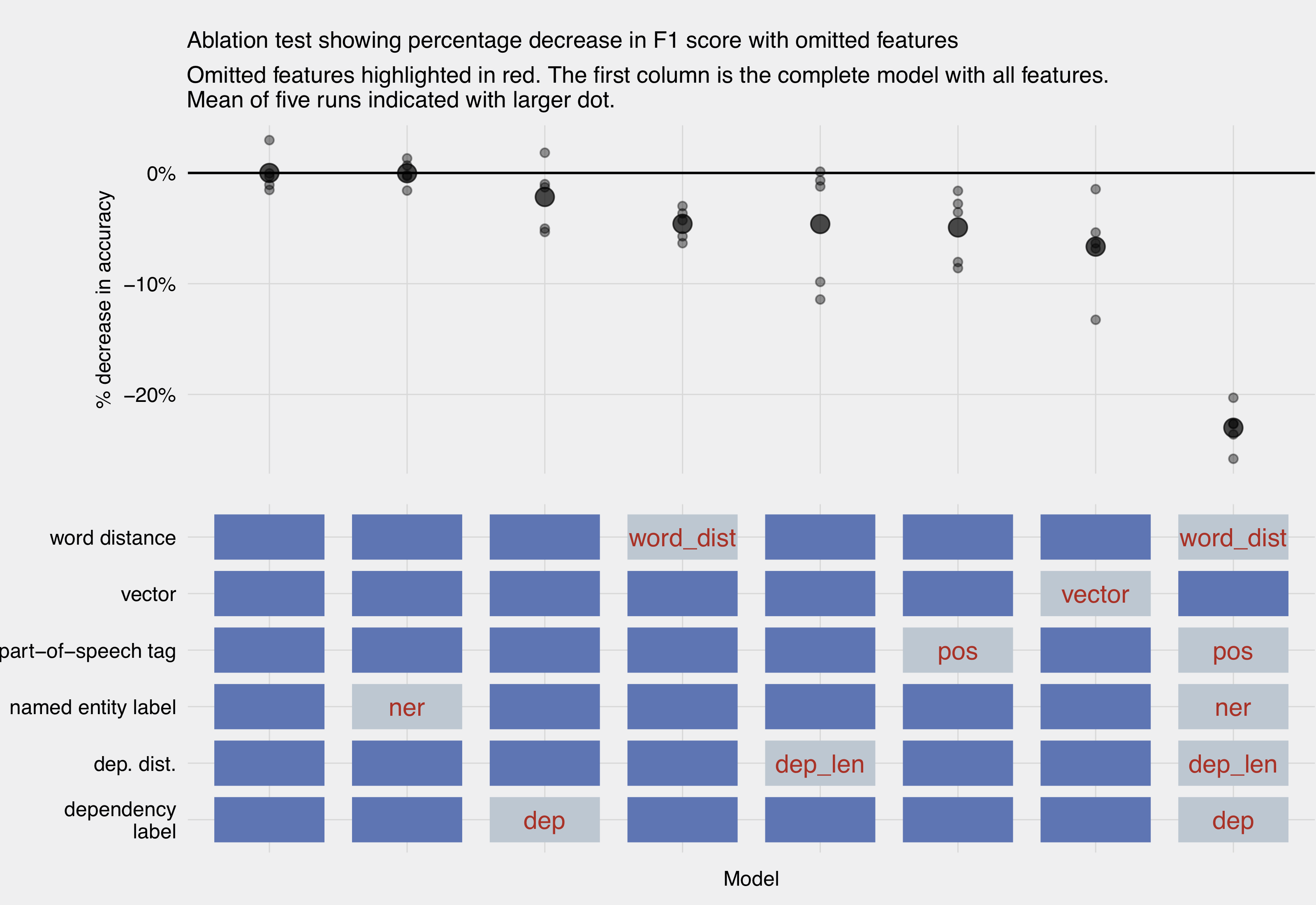}
\caption{Ablation test showing decrease in F1 score with omitted features on a
test set. Full model includes dependency labels, pre-trained GloVe embeddings,
part-of-speech tags, named entity labels, the (signed) distance from the word
to the verb, and the length of the dependency path from the word to the verb.
All conditions used the same neural net model, with the best performing model
on a validation set applied to held out test set.}
\label{fig:abl}
\end{figure*}

Figure \ref{fig:abl} shows the results of an ablation process on the
best performing LSTM model, revealing that some features are more
important than others across several random partitions and retrainings
of the model.

The ablation test reveals several interesting findings. First, the
variability in feature importance across different train-test splits of
the data prevents overly strong claims. With that in mind, the named
entity label returned by spaCy would seem to be a useful feature in a
task that requires picking one of potentially several place names. In
fact, removing it leaves the accuracy unchanged, perhaps because the
labeled data skews toward Arabic place names, which spaCy's model
struggles to recover. The two distance features, one encoding distance
from each word to the verb of interest and the other encoding the length
of the shortest dependency path between them, both seem marginally
helpful. Surprisingly, the part-of-speech feature is more useful than
the dependency label. This may be because the tree structure of the
dependency parse is not being incorporated, only its labels. Finally,
the pretrained GloVe embedding feature is helpful (second to the right
column), but it is by no means sufficient on its own (rightmost column).
While some of the literature on neural networks for NLP simply starts
from pretrained word or character embeddings and learns useful
representations from those, these results indicate that wider feature
inclusion is very helpful for the model's accuracy. The result is not
driven solely by place names being out-of-vocabulary, as GloVe contains
embeddings for 78\% of the place names in the corpus.

Qualitative inspection of miscoded sentences also reveals that the model
often fails to select the more specific location when one is available.
Performing the geoparsing step first, and then incorporating that
information into the event linking step could reduce this mode of
failure. Future work could also replace categorical features, such as
POS and dependency labels, with embeddings (see, e.g. Nguyen and
Grishman 2015).

\section{Application: Geolocating offensives in
Syria}\label{application-geolocating-offensives-in-syria}

To demonstrate the usefulness of this approach, I use it to create a
dataset of Syrian military offensives in 2016 by automatically coding
military offensive events from text and geolocating them.

I collected 15,000 news stories on Syria covering 2016 from four
sources: Al-Masdar news, Middle East Eye, Ara News, and news put out by
the opposition National Coalition. To recognize the events themselves in
the text, I created a one-off event coder that performs a dependency
parse of the documents in the corpus and compares different grammatical
parts of the sentence with a hand-specified set of terms to describe
military offensives.

After recognizing an event in the text, I then use my event geoparsing
method to find the location(s) in the text linked to the event's verb.
In order to produce final usable event data, I also perform the final
step of resolving the event location or locations to their geographical
coordinates. To do so I use the Mordecai text geoparser (Halterman
2017), which uses a neural network trained on several thousand
gold-standard resolved place names to infer the country of a location
mention, then performs a fuzzy-string search over the Geonames gazetteer
(Wick and Boutreux 2011) and selects the best location among the
locations returned from the search.

\begin{figure}
\caption{Locations of reported offensives in Syria in 2016}
\centering
\label{fig:map}
\includegraphics[width=1\columnwidth]{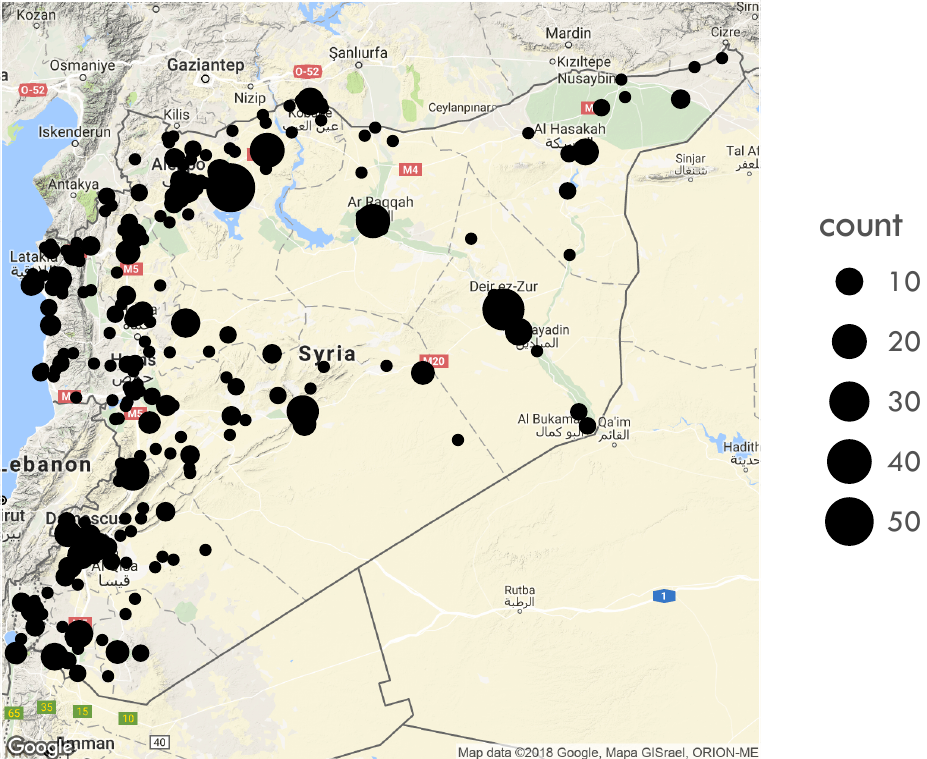}
\end{figure}

When combined with geolocated data on civilian deaths in Syria
(Halterman 2018), the geolocated offensives allow us to determine that
around 7\% of civilian deaths in Syria occurred within one day and 1
kilometer of an announced military operation. This new dataset
contributes to a growing literature on violence against civilians in
civil war, showing that even in a conventional civil war like Syria's,
only a relatively small number of casualties are plausibly related to
collateral damage from military operations. Figure \ref{fig:map} shows
the geographic distribution of new offensives. This ability to create a
dataset of when and where conventional fighting is occurring paves the
way for better understanding of the patterns of violence against
civilians in civil wars.

\section{Conclusion}\label{conclusion}

This paper introduces a state-of-the-art technique for linking events
and locations in text with performance as good as humans. It proposes a
new conceptualization of this task, focusing more on broad applicability
than previous approaches in natural language processing, but more
carefully accounting for grammar and the potential multiplicity of
events than previous work in political science. It introduces a new
labeled corpus of events and their locations, making the task accessible
to other NLP researchers. The trained model achieves an F1 score of
0.82, making it accurate enough for researchers to begin to use.

In the social sciences, the availability of a model that can link events
and locations in text should greatly increase the utility of event-type
data for subnational researchers. Event data research on police violence
in the United States (Keith et al. 2017), protest mobilization (Hanna
2017), political violence (Hammond and Weidmann 2014), and instability
forecasting (Ward et al. 2013) could all be greatly improved by better
techniques for automatically geolocating events. Researchers'
understandings of many of these social phenomena are limited by the
availability of very fine-grained geographic data.

Future NLP work could improve accuracy by integrating the ``toponym
resolution'' and event--location linking steps to improve accuracy, and
could extend the model beyond a single sentence to increase the range of
event types that the method can be applied to.

More broadly, this work builds on a growing body of research at the
intersection of NLP and social science that attempts to extract
information from text, rather than summarizing or categorizing
documents. Text also holds a great deal of factual information and new
techniques are needed to allow researchers to extract political
information from text. The technique introduced here will improve
researchers' ability to incorporate information extracted from text into
research studies that rely on geographically fine-grained data.

\section{Acknowledgements}\label{acknowledgements}

For valuable feedback on stages of this paper, I thank John Beieler,
Fotini Christia, In Song Kim, Rich Nielsen, David Smith, Brandon
Stewart, Rachel Tecott, and two very helpful anonymous reviewers. Emily
Young provided excellent assistance in annotating training data. I
gratefully acknowledge the support of a National Science Foundation
Graduate Research Fellowship. For support in developing Mordecai
(https://github.com/openeventdata/mordecai) and for creating annotated
text data, I thank the National Science Foundation under award number
SBE-SMA-1539302, the Defense Advanced Research Project Agency's XDATA
program, and the U.S. Army Research Laboratory and the U.S. Army
Research Office through the Minerva Initiative under grant number
W911NF-13-0332. Any opinions, findings, and conclusions or
recommendations expressed in this material are those of the author and
do not necessarily reflect the views of the National Science Foundation
or the Department of Defense.

\section{References}\label{references}

\hypertarget{refs}{}
\hypertarget{ref-aone2000rees}{}
Aone, Chinatsu, and Mila Ramos-Santacruz. 2000. ``REES: A Large-Scale
Relation and Event Extraction System.'' In \emph{Proceedings of the
Sixth Conference on Applied Natural Language Processing}, 76--83.
Association for Computational Linguistics.

\hypertarget{ref-avvenutigsp2018gsp}{}
Avvenuti, Marco, Stefano Cresci, Leonardo Nizzoli, and Maurizio Tesconi.
2018. ``GSP (Geo-Semantic-Parsing): Geoparsing and Geotagging with
Machine Learning on Top of Linked Data.'' \emph{Extended Semantic Web
Conference (ESWC)}.

\hypertarget{ref-bai2018empirical}{}
Bai, Shaojie, J. Zico Kolter, and Vladlen Koltun. 2018. ``An Empirical
Evaluation of Generic Convolutional and Recurrent Networks for Sequence
Modeling.'' \emph{arXiv Preprint arXiv:1803.01271}.

\hypertarget{ref-baker1998berkeley}{}
Baker, Collin F, Charles J Fillmore, and John B Lowe. 1998. ``The
Berkeley FrameNet Project.'' In \emph{Proceedings of the 17th
International Conference on Computational Linguistics-Volume 1}, 86--90.
Association for Computational Linguistics.

\hypertarget{ref-beieler2016arxiv}{}
Beieler, John. 2016. ``Generating Politically-Relevant Event Data.''
\emph{CoRR}. \url{http://arxiv.org/abs/1609.06239}.

\hypertarget{ref-beieler2016generating}{}
Beieler, John, Patrick T Brandt, Andrew Halterman, Erin Simpson, and
Philip A Schrodt. 2016. ``Generating Political Event Data in Near Real
Time: Opportunities and Challenges.'' In \emph{Data Analytics in Social
Science, Government, and Industry}, edited by R. Michael Alvarez.
Cambridge University Press.

\hypertarget{ref-berico2013clavin}{}
Berico Technologies. n.d. ``CLAVIN: Cartographic Location and Vicinity
Indexer.''

\hypertarget{ref-icews2015}{}
Boschee, Elizabeth, Jennifer Lautenschlager, Sean O'Brien, Stephen M
Shellman, James Starz, and Michael D Ward. 2015. ``ICEWS Coded Event
Data.'' In \emph{Harvard Dataverse, V9,
http://dx.doi.org/10.7910/DVN/28075}.

\hypertarget{ref-chung2017inferring}{}
Chung, Jin-Woo, Wonsuk Yang, Jinseon You, and Jong C Park. 2017.
``Inferring Implicit Event Locations from Context with Distributional
Similarities.'' In \emph{IJCAI}, 979--85.

\hypertarget{ref-doddington2004automatic}{}
Doddington, George R, Alexis Mitchell, Mark A Przybocki, Lance A
Ramshaw, Stephanie Strassel, and Ralph M Weischedel. 2004. ``The
Automatic Content Extraction (ACE) Program-Tasks, Data, and
Evaluation.'' In \emph{LREC}, 2:1.

\hypertarget{ref-douglass2018measuring}{}
Douglass, Rex W, and Kristen A Harkness. 2018. ``Measuring the Landscape
of Civil War: Evaluating Geographic Coding Decisions with Historic Data
from the Mau Mau Rebellion.'' \emph{Journal of Peace Research}.

\hypertarget{ref-d2014cliff}{}
D'Ignazio, Catherine, Rahul Bhargava, Ethan Zuckerman, and Luisa Beck.
2014. ``CLIFF-CLAVIN: Determining Geographic Focus for News.''
\emph{NewsKDD: Data Science for News Publishing, at KDD} 2014.

\hypertarget{ref-goldberg2017neural}{}
Goldberg, Yoav. 2017. \emph{Neural Network Methods for Natural Language
Processing}. Synthesis Lectures on Human Language Technologies. Morgan
\& Claypool Publishers.

\hypertarget{ref-gritta2017missing}{}
Gritta, Milan, Mohammad Taher Pilehvar, Nut Limsopatham, and Nigel
Collier. 2017. ``What's Missing in Geographical Parsing?''
\emph{Language Resources and Evaluation}. Springer, 1--21.

\hypertarget{ref-halterman2017mordecai}{}
Halterman, Andrew. 2017. ``Mordecai: Full Text Geoparsing and Event
Geocoding.'' \emph{The Journal of Open Source Software} 2 (9).
doi:\href{https://doi.org/10.21105/joss.00091}{10.21105/joss.00091}.

\hypertarget{ref-halterman2018violence}{}
---------. 2018. ``Violence Against Civilians in Syria's Civil War.''
\emph{MIT Political Science Department Research Paper}.

\hypertarget{ref-hammond2014using}{}
Hammond, Jesse, and Nils B Weidmann. 2014. ``Using Machine-Coded Event
Data for the Micro-Level Study of Political Violence.'' \emph{Research
\& Politics} 1 (2).

\hypertarget{ref-hanna2017mpeds}{}
Hanna, Alex. 2017. ``MPEDS: Automating the Generation of Protest Event
Data.'' \emph{SocArXiv Https://Osf. Io/Preprints/Socarxiv/Xuqmv}.

\hypertarget{ref-he2016deep}{}
He, Kaiming, Xiangyu Zhang, Shaoqing Ren, and Jian Sun. 2016. ``Deep
Residual Learning for Image Recognition.'' In \emph{Proceedings of the
Ieee Conference on Computer Vision and Pattern Recognition}, 770--78.

\hypertarget{ref-hill2009georeferencing}{}
Hill, Linda L. 2009. \emph{Georeferencing: The Geographic Associations
of Information}. MIT Press.

\hypertarget{ref-hochreiter1997long}{}
Hochreiter, Sepp, and Jürgen Schmidhuber. 1997. ``Long Short-Term
Memory.'' \emph{Neural Computation} 9 (8). MIT Press: 1735--80.

\hypertarget{ref-honnibal2017spacy}{}
Honnibal, Matthew, and Ines Montani. 2017. ``SpaCy 2: Natural Language
Understanding with Bloom Embeddings, Convolutional Neural Networks and
Incremental Parsing.'' \emph{To Appear}.

\hypertarget{ref-imani2017focus}{}
Imani, Maryam Bahojb, Swarup Chandra, Samuel Ma, Latifur Khan, and
Bhavani Thuraisingham. 2017. ``Focus Location Extraction from Political
News Reports with Bias Correction.'' In \emph{Big Data (Big Data), 2017
IEEE International Conference on}, 1956--64. IEEE.

\hypertarget{ref-jurafsky2018speech}{}
Jurafsky, Dan, and James H Martin. 2018. \emph{Speech and Language
Processing}. 3rd ed. draft. draft:
https://web.stanford.edu/~jurafsky/slp3/.

\hypertarget{ref-kalyvas2004urban}{}
Kalyvas, Stathis N. 2004. ``The Urban Bias in Research on Civil Wars.''
\emph{Security Studies} 13 (3). Taylor \& Francis: 160--90.

\hypertarget{ref-keith2017identifying}{}
Keith, Katherine A, Abram Handler, Michael Pinkham, Cara Magliozzi,
Joshua McDuffie, and Brendan O'Connor. 2017. ``Identifying Civilians
Killed by Police with Distantly Supervised Entity-Event Extraction.''
\emph{arXiv Preprint arXiv:1707.07086}.

\hypertarget{ref-lautenschlager2017statistical}{}
Lautenschlager, Jennifer, James Starz, and Ian Warfield. 2017. ``A
Statistical Approach to the Subnational Geolocation of Event Data.'' In
\emph{Advances in Cross-Cultural Decision Making}, 333--43. Springer.

\hypertarget{ref-lee2018lost}{}
Lee, Sophie J, Howard Liu, and Michael D Ward. 2018. ``Lost in Space:
Geolocation in Event Data.'' \emph{Political Science Research and
Methods}, 1--18.

\hypertarget{ref-leidner2008toponym}{}
Leidner, Jochen L. 2008. \emph{Toponym Resolution in Text: Annotation,
Evaluation and Applications of Spatial Grounding of Place Names}.
Universal-Publishers.

\hypertarget{ref-marcheggiani2017encoding}{}
Marcheggiani, Diego, and Ivan Titov. 2017. ``Encoding Sentences with
Graph Convolutional Networks for Semantic Role Labeling.'' \emph{arXiv
Preprint arXiv:1703.04826}.

\hypertarget{ref-nguyen2015relation}{}
Nguyen, Thien Huu, and Ralph Grishman. 2015. ``Relation Extraction:
Perspective from Convolutional Neural Networks.'' In \emph{Proceedings
of the 1st Workshop on Vector Space Modeling for Natural Language
Processing}, 39--48.

\hypertarget{ref-oconnor2013learning}{}
O'Connor, Brendan, Brandon Stewart, and Noah A Smith. 2013. ``Learning
to Extract International Relations from Political Context.''
\emph{Proceedings of the 51st Annual Meeting of the Association for
Computational Linguistics (Volume 1: Long Papers)} Vol. 1.

\hypertarget{ref-palmer2005proposition}{}
Palmer, Martha, Daniel Gildea, and Paul Kingsbury. 2005. ``The
Proposition Bank: An Annotated Corpus of Semantic Roles.''
\emph{Computational Linguistics} 31 (1). MIT Press: 71--106.

\hypertarget{ref-pustejovsky2015semeval}{}
Pustejovsky, James, Parisa Kordjamshidi, Marie-Francine Moens, Aaron
Levine, Seth Dworman, and Zachary Yocum. 2015. ``SemEval-2015 Task 8:
SpaceEval.'' In \emph{Proceedings of the 9th International Workshop on
Semantic Evaluation (Semeval 2015)}, 884--94.

\hypertarget{ref-schrodt2014three}{}
Schrodt, Philip A, John Beieler, and Muhammed Idris. 2014. ``Three's a
Charm?: Open Event Data Coding with \textsc{EL:DIABLO},
\textsc{PETRARCH}, and the Open Event Data Alliance.''

\hypertarget{ref-schrodt1994political}{}
Schrodt, Philip A, Shannon G Davis, and Judith L Weddle. 1994.
``Political Science: KEDS---a Program for the Machine Coding of Event
Data.'' \emph{Social Science Computer Review} 12 (4): 561--87.

\hypertarget{ref-speriosu2013text}{}
Speriosu, Michael, and Jason Baldridge. 2013. ``Text-Driven Toponym
Resolution Using Indirect Supervision.'' In \emph{ACL (1)}, 1466--76.

\hypertarget{ref-strotgen2016domain}{}
Strötgen, Jannik, and Michael Gertz. 2016. ``Domain-Sensitive Temporal
Tagging.'' \emph{Synthesis Lectures on Human Language Technologies} 9
(3). Morgan \& Claypool Publishers: 1--151.

\hypertarget{ref-ward2013learning}{}
Ward, Michael D, Nils W Metternich, Cassy L Dorff, Max Gallop, Florian M
Hollenbach, Anna Schultz, and Simon Weschle. 2013. ``Learning from the
Past and Stepping into the Future: Toward a New Generation of Conflict
Prediction.'' \emph{International Studies Review} 15 (4): 473--90.

\hypertarget{ref-white2016universal}{}
White, Aaron Steven, Drew Reisinger, Keisuke Sakaguchi, Tim Vieira,
Sheng Zhang, Rachel Rudinger, Kyle Rawlins, and Benjamin Van Durme.
2016. ``Universal Decompositional Semantics on Universal Dependencies.''
In \emph{Proceedings of the 2016 Conference on Empirical Methods in
Natural Language Processing}, 1713--23. Austin, Texas: Association for
Computational Linguistics.

\hypertarget{ref-wick2011geonames}{}
Wick, Marc, and C Boutreux. 2011. ``GeoNames.'' \emph{GeoNames
Geographical Database}.

\end{document}